\documentclass[11pt]{article}

\usepackage[final]{acl}

\usepackage{times}
\usepackage{latexsym}

\usepackage[T1]{fontenc}

\usepackage[utf8]{inputenc}

\usepackage{microtype}

\usepackage{inconsolata}

\usepackage{graphicx}
\usepackage{booktabs}
\usepackage{multirow}
\usepackage{graphicx}
\usepackage[table, dvipsnames]{xcolor}
\usepackage{enumitem}
\usepackage{multirow, makecell}
\usepackage{amsmath, amssymb}
\usepackage{wrapfig}
\usepackage{rotating}
\definecolor{light-gray}{gray}{0.9}

\title{Chain-of-Thought Degrades Visual Spatial Reasoning Capabilities of Multimodal LLMs}

\renewcommand{\thefootnote}{\fnsymbol{footnote}}
\author{
\begin{tabular}[t]{c@{\qquad\qquad}c}
Sai Srinivas Kancheti\footnotemark[1]\footnotemark[2] & Aditya Sanjiv Kanade\footnotemark[1]\\
IIT, Hyderabad & Microsoft Research India\\
{\tt\small cs21resch01004@iith.ac.in} & {\tt\small kanade850@gmail.com}\\[1em]
 Vineeth N. Balasubramanian & Tanuja Ganu \\
Microsoft Research India & Microsoft Research India\\
{\tt\small vineeth.nb@microsoft.com} & {\tt\small tanuja.ganu@microsoft.com}\\
\end{tabular}
}   

\begin{document}
\maketitle
\footnotetext[1]{Equal contribution.}
\footnotetext[2]{Work done while at Microsoft Research India.}
\renewcommand{\thefootnote}{\arabic{footnote}}
\begin{abstract}
Multimodal Reasoning Models (MRMs) leveraging Chain-of-Thought (CoT) based thinking have revolutionized mathematical and logical problem-solving. However, we show that this paradigm struggles with generalized spatial intelligence. We perform a comprehensive evaluation of seventeen models across thirteen spatial benchmarks and identify a critical gap: CoT prompting consistently degrades performance in visual spatial reasoning. 
Furthermore, through a novel \textit{No-Image++} ablation, we demonstrate that MRMs and CoT prompted MLMs suffer from severe shortcut learning, and hallucinate visual details from textual priors even when the image is absent. These findings challenge the efficacy of text-only CoT for spatial tasks and underscore the need for vision-centric reasoning paradigms.

\end{abstract}

\section{Introduction}

The emergence of "System 2" Multimodal Reasoning Models (MRMs)\,---\,models post-trained via SFT and RL to generate step-by-step reasoning\,---\,has driven remarkable progress in mathematical and logical domains. By leveraging Reinforcement Learning (RL)~\cite{TULU3,DeepSeekAI2025DeepSeekR1IR} and long Chain-of-Thought (CoT)~\cite{cot1,cot2} inference, MRMs demonstrate the ability to self-correct and reason through complex problems. Separately, \emph{CoT prompting} is a general technique that instructs any Multimodal Language Model (MLM) to think step-by-step before answering. However, a fundamental question remains: \textbf{does this text-centric reasoning paradigm translate to spatial intelligence?} Spatial reasoning requires grounding, geometric intuition, and precise localization, which are skills that may not easily arise from verbose, text-based reasoning~\cite{cambrian1,MMVP}.

\begin{figure*}[t]
    \centering
    \begin{minipage}[t]{0.3\linewidth}\vspace{0pt}
    \centering
    \scalebox{0.76}{
    \begin{tabular}{l|cc}
    \hline
    Model & CoT & Non-CoT  \\
    \hline
    GThinker   & 62.52 & 39.38 ($\textbf{-23.14}\%$) \\
    R1-Ov      & 46.88 & 47.84 ($\textbf{+0.96}\%$) \\
    ViGoRL     & 60.68 & 62.52 ($\textbf{+1.84}\%$) \\
    VL-Re.     & 60.99 & 62.18 ($\textbf{+1.19}\%$) \\
    Vision-G1  & 63.26 & 62.85 ($\textbf{-0.41}\%$) \\
    Vision-R1  & 58.86 & 59.6  ($\textbf{+0.74}\%$) \\
    TreeVGR    & 61.11 & 62.6  ($\textbf{+1.49}\%$) \\
    ThinkLite  & 62.61 & 62.74 ($\textbf{+0.13}\%$) \\
    \hline
    \end{tabular}
    }
    \end{minipage}\hfill
    \begin{minipage}[t]{0.68\linewidth}\vspace{0pt}
    \centering
    \includegraphics[width=\linewidth]{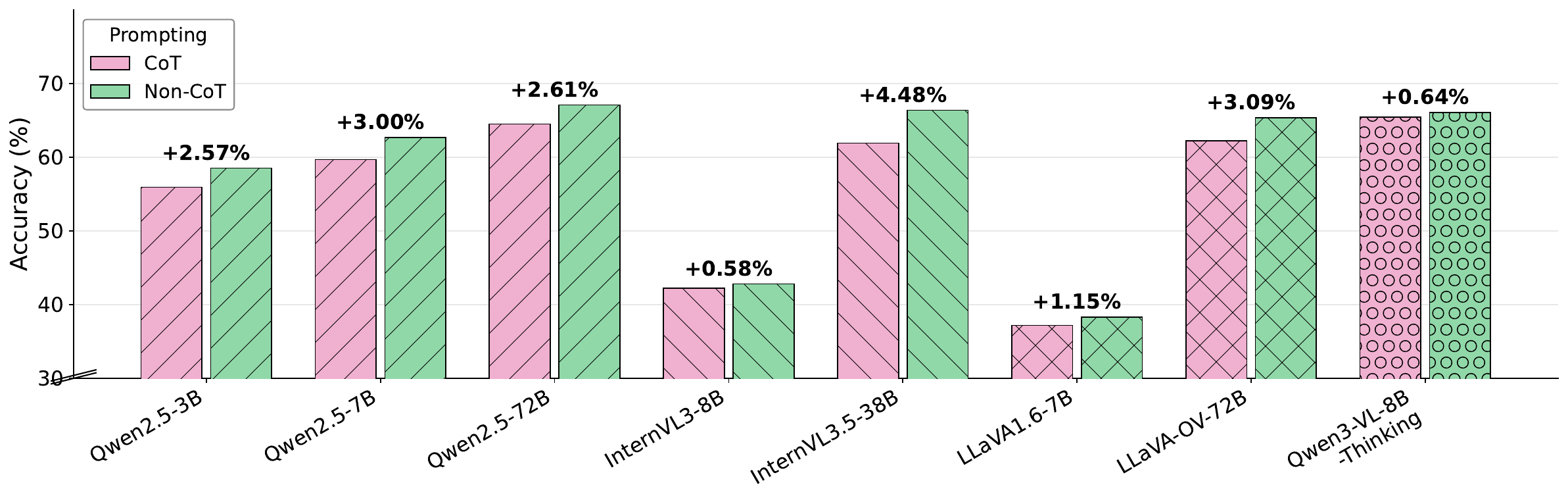}
    \end{minipage}
    \caption{(Left) CoT vs Non-CoT performance of open-source MRMs. (Right) Bar chart showing the average accuracy of various families of MLMs over 13 benchmark datasets. For each model, the left bar shows the accuracy achieved by CoT prompting, and the right bar shows for base prompt (non-CoT). We observe CoT prompting drops performance over a wide range of backbones and model scales, including Qwen3-VL-8B-Thinking~\cite{qwen3vl}, a model with explicitly enhanced spatial perception.}
    \label{fig:cot_vs_noncot_mlm_bar}
\end{figure*}

In this work, we conduct a comprehensive evaluation of seventeen models, including nine state-of-the-art open-source MRMs (e.g., GThinker, Vision-R1, ViGoRL, Qwen3-VL) and eight diverse backbone MLMs. We benchmark these models across thirteen datasets covering static 2D relations, 3D geometry, and dynamic/temporal understanding. To isolate the impact of CoT reasoning, we standardize our evaluation using a uniform evaluation and scoring policy. Our findings reveal that contrary to trends in other domains, CoT prompting degrades performance in visual spatial tasks. Our contributions are as follows: (i) We show that MRMs consistently underperform their own backbone on generalized spatial benchmarks. In our experiments, 7 out of 8 reasoning models failed to surpass the backbone they were distilled from. (ii) We demonstrate in Figure~\ref{fig:cot_vs_noncot_mlm_bar} that CoT prompting lowers accuracy by an average of 3\% across a diverse range of MLMs. (iii) Through a novel \textit{No-Image++} ablation, we show that MRMs suffer from severe shortcut learning. When presented with a blank image and a ``Cannot determine'' option, reasoning models continue to hallucinate visual details and confidently select incorrect answers based solely on textual priors.

These results suggest that simply scaling text-based reasoning is insufficient for robust spatial intelligence, highlighting the need for vision-centric training paradigms.

\begin{table*}[htb]
    \centering
    \scalebox{0.8}{
    \begin{tabular}{l|ccccccc}
    \toprule
    Models & 3DSRBench & BLINK & \multicolumn{2}{c}{CV-Bench} & MindCube & MMSIBench & MMVP \\
    \cmidrule(lr){4-5}&&& 2D & 3D && \\
    \midrule
    Qwen2.5-VL-7B$_{cot}$ & $\textbf{57.11}_{0.39}$& $53.44_{0.30}$& $75.92_{0.03}$& $76.09_{0.31}$& $30.83_{0.25}$& $\textbf{27.47}_{0.21}$& $72.44_{0.68}$ \\
    Qwen2.5-VL-7B & $55.38_{0.06}$& $\textbf{56.04}_{0.03}$& $\underline{77.17}_{0.03}$& $\underline{83.78}_{0.04}$& $35.11_{0.18}$& $26.87_{0.09}$& $\underline{75.78}_{0.32}$ \\
    \midrule
    \rowcolor{light-gray}
    GThinker-7B & $56.58_{0.20}$& $54.76_{0.17}$& $\textbf{77.40}_{0.06}$& $82.95_{0.04}$& $\underline{40.16}_{0.32}$& $\underline{27.33}_{0.31}$& $73.78_{0.42}$ \\
    R1-Onevision-7B & $48.52_{0.20}$& $43.27_{0.50}$& $53.31_{0.04}$& $58.00_{0.42}$& $27.09_{0.42}$& $13.30_{0.30}$& $56.16_{0.16}$ \\
    \rowcolor{light-gray}
    ViGoRL-7B-Spatial & $55.84_{0.20}$& $52.51_{0.26}$& $76.59_{0.27}$& $\textbf{86.14}_{0.10}$& $39.36_{0.16}$& $25.87_{0.17}$& $73.22_{0.42}$ \\
    VL-Rethinker-7B & $\underline{56.99}_{0.11}$& $\underline{54.60}_{0.23}$& $76.06_{0.12}$& $80.75_{0.14}$& $37.81_{0.27}$& $26.90_{0.08}$& $75.89_{0.16}$ \\
    \rowcolor{light-gray}
    Vision-G1 & $55.91_{0.01}$& $\underline{54.60}_{0.08}$& $76.70_{0.15}$& $\underline{83.75}_{0.14}$& $38.10_{0.31}$& $26.07_{0.12}$& $\textbf{76.56}_{0.16}$ \\
    Vision-R1-7B & $55.01_{0.20}$& $46.47_{0.66}$& $71.58_{0.12}$& $75.83_{0.31}$& $36.95_{0.59}$& $22.90_{0.36}$& $72.22_{0.32}$  \\
    \rowcolor{light-gray}
    TreeVGR-7B & $51.53_{0.03}$& $53.16_{0.25}$& $76.24_{0.09}$& $75.17_{0.13}$& $\textbf{44.25}_{0.66}$& $27.17_{0.33}$ & $71.33_{0.27}$ \\
    ThinkLite-7B & $57.26_{0.13}$& $57.13_{0.08}$& $76.89_{0.18}$& $80.44_{0.16}$& $30.13_{0.23}$& $27.97_{0.56}$& $73.56_{0.16}$ \\
    \end{tabular}
    }
    \vspace{1em}
    \scalebox{0.8}{
    \begin{tabular}{l|cccccc|c}
    \toprule
    Models & OmniSpatial & RealWorldQA & SAT & SpatialBench & VSR & V*Bench & Avg. \\
    \midrule
    Qwen2.5-VL-7B$_{cot}$ & $40.40_{0.80}$& $63.05_{0.27}$& $59.22_{0.57}$& $61.75_{0.70}$& $81.83_{0.35}$& $76.27_{0.25}$& 59.68 \\
    Qwen2.5-VL-7B & $45.23_{0.11}$& $\underline{69.02}_{0.00}$& $\underline{63.11}_{0.16}$& $\underline{62.87}_{0.00}$& $\underline{85.38}_{0.04}$& $79.06_{0.00}$& \underline{62.68} \\
    \midrule
    \rowcolor{light-gray}
    GThinker-7B & $\textbf{47.68}_{0.14}$& $68.67_{0.06}$& $58.44_{0.16}$& $60.07_{0.15}$& $83.77_{0.04}$& $\underline{81.15}_{0.00}$& 62.52 \\
    R1-Onevision-7B & $31.54_{0.10}$& $49.87_{0.46}$& $51.50_{0.83}$& $50.19_{0.18}$& $72.50_{0.32}$& $54.19_{0.79}$& 46.88 \\
    \rowcolor{light-gray}
    ViGoRL-7B-Spatial & $36.97_{0.21}$& $65.67_{0.55}$& $58.44_{0.68}$& $58.65_{0.49}$& $82.08_{0.07}$& $77.49_{0.74}$& 60.68 \\
    VL-Rethinker-7B & $39.84_{0.26}$& $68.50_{0.39}$& $\textbf{65.00}_{0.98}$& $\underline{61.57}_{0.15}$& $\underline{84.40}_{0.08}$& $64.57_{1.08}$& 60.99 \\
    \rowcolor{light-gray}
    Vision-G1 & $46.88_{0.21}$& $\textbf{69.76}_{0.16}$& $62.67_{0.00}$& $\textbf{64.93}_{0.00}$& $\textbf{86.55}_{0.08}$& $79.93_{0.25}$& \textbf{63.26} \\
    Vision-R1 & $39.75_{0.16}$& $67.41_{0.06}$& $58.45_{0.32}$& $60.51_{0.08}$& $79.95_{0.18}$& $78.18_{0.25}$& 58.86 \\
    \rowcolor{light-gray}
    TreeVGR-7B & $\underline{47.29}_{0.38}$& $67.58_{0.11}$& $62.11_{0.79}$& $60.26_{0.16}$& $74.80_{0.07}$& $\textbf{83.60}_{0.49}$& 61.11 \\
    ThinkLite-7B & $45.36_{0.43}$& $69.37_{0.06}$& $66.44_{0.42}$& $62.19_{0.32}$& $86.85_{0.04}$& $80.28_{0.25}$& 62.61 \\
    \end{tabular}
    }
    \caption{Accuracy of SOTA MRMs on 13 spatial benchmarks. The top two rows shows performance of the base model Qwen2.5-VL-7B, which is competitive with MRMs trained to perform multimodal reasoning. We identify that open-source MRMs do not exhibit generalized spatial intelligence beyond their base model.}
    \label{tab:main_table}
\end{table*}

\section{Methodology}
\label{sec:methodology}
We begin by describing the baseline models we consider, datasets and our evaluation scheme.

\noindent\textbf{Baselines:} We rigorously benchmark the performance of seventeen models:

\noindent\textbf{i) Qwen2.5VL backbone}: We evaluate three scales of Qwen2.5-VL-Instruct series (3B, 7B, 72B)~\cite{qwen2.5vl_techreport}, \underline{eight} SOTA General purpose visual reasoning Multimodal Reasoning Models (MRMs) trained using RLVR~\cite{grpo} GThinker-7B~\cite{zhan2025gthinker} (Jun'25), {ViGoRL-7B-Spatial}~\cite{vigorl} (May'25), {Vision-G1-7B}~\cite{Vision-g1} (Aug'25), {R1-Onevision-7B}~\cite{Yang2025R1OnevisionAG} (Mar'25), {VL-Rethinker-7B}~\cite{Wang2025VLRethinkerIS} (May'25), {Vision-R1}~\cite{Huang2025VisionR1IR} (Mar'25), TreeVGR~\cite{treevgr} (Jul'25), and ThinkLite-7B~\cite{thinklite} (Apr'25); Qwen3-VL-8B-Thinking~\cite{qwen3vl}, a model with explicitly enhanced spatial perception; \textbf{ii) InternVL backbone:} InternVL3-8B, InternVL3.5-38B~\cite{internvl3.5}; \textbf{iii) LLaVA backbone:} LLaVA-v1.6-Mistral-7B, LLaVA-OneVision-Qwen2-72B~\cite{llavaov}; and finally a proprietary model GPT-4o~\cite{Hurst2024GPT4oSC} for a total of \emph{seventeen} diverse baselines covering both reasoning and non-reasoning MLMs at various scales.

\noindent\textbf{Open-source MRMs.} We choose eight diverse top performing open-source MRMs (listed above) that are trained for general visual reasoning including spatial tasks. ViGoRL-spatial and TreeVGR are explicitly trained to perform spatial reasoning, while the remaining MRMs (with the exception of Vision-R1) contain spatial domains in their training data and are trained as general purpose visual reasoners. All MRMs considered are built atop the Qwen2.5-VL-7B-Instruct backbone, which combines strong visual capabilities with emergent reasoning abilities, and has been widely used as a capable backbone.
A key motivation for our study, highlighted in Tab.~\ref{tab:baseline_eval_summary}, is that the original papers for these models evaluate on \emph{math heavy} datasets which are \emph{not vision-centric}. We hence comprehensively evaluate the reasoning capabilities of these models on spatial tasks.\\


\begin{table}[htb]
    \centering
    \small

    \parbox[t]{0.48\textwidth}{
    \centering
    \resizebox{\linewidth}{!}{
    \begin{tabular}{l|c}
    \toprule
    Baseline & Paper-Reported Datasets \\
    \midrule
    GThinker-7B & MMStar, RWQA, MMMU-Pro\\
    R1-Onevision-7B &  MathVision, Mathvista, Mathverse\\
    ViGoRL-7B-Spatial & SAT-Val, BLINK \\
    VL-Rethinker & MathVision, MMMU-Pro, MEGA\\
    Vision-G1 & MathVista, MMMU-Pro, MMStar, ChartQA \\
    Vision-R1 & MathVista, MMStar, ChartQA, MME$_{sum}$\\
    \end{tabular}
    }
    }
    \caption{A summary of the evaluation datasets used in each MRM's paper. Most eval datasets are Math heavy, are not vision-centric, and do not cover many aspects of spatial reasoning.}
    \label{tab:baseline_eval_summary}
\end{table}

\begin{table*}[htb]
    \centering
    \small
    \scalebox{0.58}{
    {\Large
    \setlength{\tabcolsep}{3pt}
    \begin{tabular}{l|c|cc|cccccccc}
    \toprule
    $\overrightarrow{\text{Models}}$ & \large{Random} & Qwen2.5$_{cot}$ & Qwen2.5 & GThinker & R1-Ov & ViGoRL & VL-Re. & Vision-G1 & Vision-R1 & TreeVGR & ThinkLite\\
    \midrule
    No-Image  & 38.83 & 37.45& 38.59& 44.17& 28.1 & 43.18 & 41.26 & 44.46 & 41.15 & 41.91& 42.48   \\
    No-Image++ & - & 43.4& \textbf{76.41}& 5.55 & 11.22 & 30.95 & 47.73 & 25.28 & 7.29 & 11.35& 36.00   \\   
    \end{tabular}}
    }
    \caption{Results of two variant of the No-Image ablation, where the images are replaced with an uninformative full gray image. For No-Image, MRMs show much higher average performance, indicating their ability to shortcut an answer just from the question (random is better). For No-Image++ (higher better), where a ``cannot determine'' option is added, we find Qwen\_cot as well as MRMs still choose other options as they are biased by the text trace.}
    \label{tab:noimg_ablation_short}
\end{table*}

\noindent\textbf{Datasets:} 
We evaluate on \underline{thirteen} datasets covering various aspects of spatial reasoning, which can be broadly categorized into static 2D datasets, and 3D/dynamic datasets. The former datasets are usually confined to single images and focus on planar spatial relationships, usually from the camera's perspective. We place BLINK~\cite{BLINK}, CV-Bench2D~\cite{cambrian1}, MMVP~\cite{MMVP}, RealWorldQA~\cite{RealWorldQA_xai}, SpatialBench~\cite{SpatialBench}, VSR~\cite{VSR}, and V*Bench~\cite{VstarBench} in this category.

We also consider benchmarks that require reasoning involving 3D geometry, depth, multi-image consistency, and temporal reasoning. These datasets often involve understanding the 3D position and relative orientation from the object's perspective inside the image. 3DSRBench~\cite{ma20243dsrbench}, CV-Bench3D~\cite{cambrian1}, MindCube~\cite{mindcube}, MMSIBench~\cite{MMSIBench}, OmniSpatial~\cite{omnispatial25}, and SAT-Real~\cite{ray2025sat} belong to this category. We choose these datasets as they have real-world objects set in natural scenes, are difficult to answer on account of being vision-centric~\cite{cambrian1}, and cover a wide rage of spatial capabilities. A summary of the datasets along with the various spatial facets they test is provided in Appx. Tab.~\ref{tab:dataset_summary}. The left half of the table describe the 2D datasets, while the right half describes the rest.\\

\noindent\textbf{Evaluation:} To ensure uniform evaluation, we follow VLMEvalKit~\cite{duan2024vlmevalkit} to provide a uniform system prompt as well as a uniform question format. All benchmarks, are multiple-choice questions (MCQs) with options provided in the question prompt. The question format for all datasets is: \texttt{Question:<question>\textbackslash nOptions:\textbackslash nA. <optA>\textbackslash nB.<optionB> ...\textbackslash nPlease select the correct answer (letter and option text) from the options above.}
We append dataset-specific prompts for OmniSpatial \& MindCube to ensure good performance. The prompts are detailed in Appx. \S~\ref{app:prompts}.

\noindent\textbf{Metric.} We use vLLM~\cite{vllm} version $0.10.0$ for performant, batched inference of MLMs on $4$ NVIDIA $A100$ GPUs. We use a batch size of $16$, set max new tokens generated to be $32768$, set model context length as $32768$, and perform inference on $bfloat16$ precision. We use pass@1 accuracy under greedy decoding (with temperature set to $0$) as our metric. All results are over 3 seeds.

\noindent\textbf{System Prompts.} We evaluate models in two settings, using a i) \emph{base prompt/non-CoT prompt} such as \texttt{``You are a spatial-reasoning assistant. The user asks a question, and the Assistant solves it.''} or a ii) \emph{CoT prompt} where we append \texttt{``First output the thinking process in <think> </think> tags and then output the final answer in <answer> </answer> tags.''} to the base prompt. For CoT evaluation of MRMs, we use the custom CoT prompt they train on (instead of default shown above) for best performance as shown in the table below.
\begin{wraptable}[3]{r}{0.32\textwidth}
\vspace{-0.5cm}
\centering
\scalebox{0.68}{
\begin{tabular}{c|c |c}
    \toprule
    Model & Custom CoT & Simple CoT \\
    \midrule
    GThinker-7B	& 62.52 & 59.57 \\
    Vision-G1 & 63.26 & 62.06  \\
    \bottomrule
\end{tabular}
}
\end{wraptable}
The custom CoT prompt used for each MRM is shown in Appx \S~\ref{app:prompts}.

\noindent\textbf{Scoring the generations.} We use an LLM-as-a-judge along with a carefully designed prompt, shown in Appx. \S~\ref{app:judge_prompts}, to score all generations. We pick Qwen3-30B-A3B-Instruct-2507, a small non-reasoning text model as our judge for scoring since our evaluation is on MCQs with short answers i.e the final answer of the model is not free-form but restricted to the options provided. To validate our choice, we re-score the generations of Vision-G1 using GPT-4o as the judge, and compare it with our chosen judge. We observe a Cohen's kappa score~\cite{cohenkappa} of $>0.99$ indicating near-perfect agreement of judges.

\section{Results and Analysis}
\noindent\textbf{(i) CoT Prompting Hurts Visual Spatial Reasoning.} Contrary to trends in math and logic domains, we observe that CoT prompting frequently hurts performance in visual spatial tasks.
Figure~\ref{fig:cot_vs_noncot_mlm_bar} (left) shows the performance of open-source MRMs under both CoT and non-CoT prompting. Surprisingly, these models, which have been explicitly trained via RL to reason, often perform better when this reasoning capability is suppressed. Six of the eight MRMs achieve higher accuracy with the non-CoT prompt than with their native CoT prompts.
We observe that GThinker is not robust to changes in its prompt and struggles to adhere to the direct-answer format of the base prompt. We qualitatively observe that it generates ill-formed CoT traces even when instructed otherwise, causing a significant drop ($-23.14\%$) in performance (see Figure~\ref{fig:qual_noimg} for an example). Figure~\ref{fig:cot_vs_noncot_mlm_bar} (right) shows that this trend holds across three families of models (Qwen, InternVL \& LLaVA), and across a range of model strengths (params ranging from 3B to 72B). We additionally evaluate Qwen3-VL-8B-Thinking~\cite{qwen3vl}, a model with explicitly enhanced spatial perception, and observe that Non-CoT still outperforms CoT by $+0.64\%$ at a competitive baseline of $\sim$$65\%$ (dataset-wise results in Appx.\ Table~\ref{tab:qwen3vl_results}).

\noindent\textbf{(ii) RL-trained Multimodal Reasoning Models Underperform their Backbone.} We present the accuracies of eight open-source MRMs on 13 datasets in Tab.~\ref{tab:main_table}. The first two rows indicate backbone results, where $cot$ indicates evaluation under CoT prompt. We observe that the non-CoT Qwen2.5-VL backbone shows strong average performance of $62.68\%$. Surprisingly, despite extensive SFT and RL training designed to enhance visual reasoning, seven out of the eight MRMs fail to surpass this baseline. Even models explicitly finetuned for spatial tasks ViGoRL ($-2\%$) \& TreeVGR ($-1.57\%$) underperform the backbone. Vision-G1 is the only exception, and outperforms the backbone by $+0.6\%$. However, we show in the next paragraph that Vision-G1 exhibits the strongest reliance on textual priors, suggesting its performance may stem partly from dataset shortcuts rather than grounded visual reasoning.



\noindent\textbf{(iii) Reasoning Models Show Over-reliance on Text Rationale.} We highlight a crucial shortcoming of text-only CoT reasoning, where we observe an over-reliance on the text modality which leads to hallucination of visual content. We perform a \emph{No-Image} ablation, where we pass an uninformative fully gray image (of the same size and aspect ratio as the original image) as input along with the question. The first row of Table~\ref{tab:noimg_ablation_short} reports the average accuracy across all 13 datasets. We observe that MRMs perform significantly better than random guessing (e.g., GThinker achieves $44.17\%$), indicating they can "shortcut" the answer by ignoring visual content and relying solely on question text, options, and world-knowledge priors.
To confirm this behavior stems from hallucination, we introduce the \emph{No-Image++} setting. Here, we maintain the gray image input but append a "Cannot determine from the image" option to the answer choices, which serves as the ground truth. As shown in the second row of Table~\ref{tab:noimg_ablation_short}, most MRMs perform poorly on this metric. Rather than acknowledging the missing visual information, they continue to hallucinate reasoning traces for the original options. We present qualitative examples of this failure mode in Figure~\ref{fig:qual_noimg}.

\begin{figure}[t]
\small
\fbox{\parbox{0.96\columnwidth}{
\textbf{Example 1: GThinker degenerate output under Non-CoT prompt}\\
\textit{Question (RWQA):} ``Which item is larger than the other?''\\[2pt]
\textit{Output:} \texttt{<tool\_call>\textbackslash n\textbackslash n\textbackslash n\textbackslash n\textbackslash n\textbackslash n\textbackslash n\textbackslash n\textbackslash n\textbackslash n\textbackslash n\textbackslash n\textbackslash n...} (repeats until termination)\\[4pt]
\rule{\columnwidth}{0.4pt}\\[2pt]
\textbf{Example 2: ViGoRL hallucination under No-Image++ (blank image)}\\
\textit{Question (CVBench2D):} ``Where is the cave located with respect to the trees?''\\[2pt]
\textit{Output:} \texttt{<think>} The cave entrance appears to be at the bottom of the image, while the trees are higher up, suggesting that the cave entrance is below the trees. (300, 450) ...The cave entrance is clearly situated at the bottom of the image, which is below the trees. (250, 550) \texttt{</think> <answer>} below \texttt{</answer>}\\[2pt]
\textit{Note:} The input image is a \textbf{blank gray image}. The model hallucinates coordinates and spatial relationships using world knowledge (caves are usually below trees).
}}
\caption{Qualitative examples of failure modes. \textbf{Top:} GThinker produces degenerate output when prompted without CoT. \textbf{Bottom:} ViGoRL hallucinates detailed spatial reasoning for a blank image in the No-Image++ setting. Additional examples in Appx.~\ref{app:qualitative}.}
\label{fig:qual_noimg}
\end{figure}

\begin{table}[t]
    \centering
    \small
    \scalebox{0.85}{
    \begin{tabular}{l|cc}
    \toprule
    Model & Avg. CoT & Avg. Non-CoT \\
    \midrule
    GPT-4o~\cite{Hurst2024GPT4oSC} & 65.55 \small{$(+0.50)$} & 65.05  \\
    GPT-4.1-mini~\cite{openai2024gpt4technicalreport} & 67.79 \small{$(+0.39)$} & 67.40  \\
    GPT-5~\cite{singh2025openaigpt5card} & 69.00  & 69.65 \small{$(+0.65)$} \\
    GPT-5-mini~\cite{singh2025openaigpt5card} & 69.86 \small{$(+0.08)$} & 69.78 \\
    GPT-5-nano~\cite{singh2025openaigpt5card} & 60.63  & 61.86 \small{$(+1.23)$} \\
    \bottomrule
    \end{tabular}
    }
    \caption{CoT vs Non-CoT performance of proprietary models. Non-CoT outperforms CoT for GPT-5 and GPT-5-nano.}
    \label{tab:proprietary}
\end{table}

\noindent\textbf{(iv) Analysis of Proprietary Models.} To evaluate the generalizability of our findings to proprietary models, we benchmark five models from the GPT family. Table~\ref{tab:proprietary} shows that non-CoT performance remains competitive with or exceeds CoT performance across the board. Notably, GPT-5 and GPT-5-nano show CoT degradation ($+0.65\%$ and $+1.23\%$ for Non-CoT, respectively), mirroring the open-source trend. While GPT-4o and GPT-4.1-mini show marginal CoT gains, these are small ($<0.5\%$) relative to the additional inference compute required for reasoning.

We analyze the CoT traces of proprietary models and find two notable differences from open-source models: (i) Proprietary models produce significantly shorter traces ($\sim$$350$ characters for GPT-5-mini vs.\ $\sim$$3600$ characters for Qwen3-VL-8B-Thinking), and (ii) proprietary traces lack the reflective phrases (e.g., ``wait'', ``let me reconsider'') and repetitive looping commonly observed in open-source MRMs. We hypothesize that this conciseness helps proprietary models avoid the hallucination-inducing verbosity that harms open-source models, though their training details remain opaque. These observations suggest that the quality and conciseness of reasoning traces, rather than their mere presence, may be key to preserving spatial reasoning performance under CoT prompting.

\section{Conclusion}
\vspace{-4pt}
In this work, we show that the success of reasoning models in logic and mathematics does not yet extend to the spatial domain. Our benchmarking of seventeen models across thirteen datasets reveals that Chain-of-Thought prompting consistently degrades spatial reasoning performance, with specialized MRMs frequently underperforming their own base models. Crucially, our \textit{No-Image++} analysis identifies the mechanism behind this failure: current reasoning chains tend to hallucinate visual information based on textual priors rather than engaging in grounded perception. Our analysis of proprietary models further suggests that concise, non-repetitive reasoning traces may mitigate this degradation. Our work highlights the need for vision-centric training paradigms for MRMs. Promising future directions include (i) test-time visual verifiers that evaluate each reasoning step against image evidence and trigger backtracking on incorrect visual claims, and (ii) visual process reward models that incentivize grounded, perception-first reasoning during training.

\section*{Limitations}

In this work, we have sought to cover a broad range of visual spatial reasoning datasets and R1‑style MRMs. However, we do not claim that the 13 datasets included represent the entirety of the visual spatial reasoning domain. Given the current landscape of MRMs, it is challenging to completely isolate all confounding factors that may lead to performance improvements or declines across these datasets. We note that proprietary model training details remain opaque, limiting deeper analysis of their behavior. We believe this study offers a solid foundation for future research to further explore vision-centric reasoning paradigms.

\section*{Acknowledgments}

\bibliography{references}

\newpage
\appendix

\section{Prompts}
\label{app:prompts}

In this section we present the dataset prompts as well as different system prompts used in the baselines.

\subsection{System Prompts}

\noindent\textbf{Base prompt.} This is the simple no-thinking prompt used by Qwen2.5-VL-7B.
\texttt{
You are a spatial-reasoning assistant. The user asks a question, and the Assistant solves it.    
}
\newline

\noindent\textbf{CoT prompts.} We give the list of CoT system prompts we use to evaluate the MRM baselines. Below we present prompts used by GThinker, R1-Onevision, ViGoRL-Spatial, VL-Rethinker, Vision-G1, and Vision-R1 respectively.

\noindent\underline{GThinker: }\texttt{
A conversation between User and Assistant. The user asks a question, and the Assistant solves it. The assistant first thinks about the reasoning process in the mind and then provides the user with the answer. \
The reasoning process and answer are enclosed within <think> </think> and <answer> </answer> tags, respectively, i.e., <think> reasoning process here </think><answer> answer here </answer>. \
In the reasoning process enclosed within <think> </think>, each specific visual cue is enclosed within <vcues\_*>...</vcues\_*>, where * indicates the index of the specific cue. \
Before concluding the final answer, pause for a quick consistency check: verify whether the visual cues support the reasoning and whether each step logically follows from what is seen. \
If correct, conclude the answer; otherwise, revise the visual cues and reasoning, then conclude.
}
\newline

\noindent\underline{R1-Onevision}: \texttt{
You are a spatial-reasoning assistant. The user asks a question, and the Assistant solves it.
First output the thinking process in <think> </think> tags and then output the final answer in <answer> </answer> tags.
}
\newline

\noindent\underline{ViGoRl-Spatial}: \texttt{
A conversation between User and Assistant. The User asks a question, and the Assistant solves it. The Assistant systematically reasons through the problem step by step by checking and verifying possible solutions and image regions, while grounding reasoning steps to specific objects and their relationships in the image using (x,y) coordinates. There may be one image or two images concatenated together, in which case the Assistant must compare the spatial relationships between the two images.\textbackslash n\textbackslash n
All reasoning processes must be enclosed within a single set of '<think>' tags, and reasoning steps must include specific reference coordinates:\textbackslash n\textbackslash n
For example, <think>
\{Reasoning text\}. \{Further reasoning text\} \{more reasoning\} 
</think>
The final answer should be enclosed in '<answer>' tags in the format:
<answer> \{text of selected answer choice\} </answer>\textbackslash n\textbackslash n
The Assistant must help the user identify the correct answer choice from the options provided.\textbackslash n
- If the correct answer is unclear, select the most relevant option based on the spatial relationships and dynamics within the image.\textbackslash n
- The Assistant should verify each step and check multiple possible solutions before selecting the final answer.
}
\newline

\noindent\underline{VL-Rethinker: }\texttt{
Please think step by step, and **regularly perform self-questioning, self-verification, self-correction to check your ongoing reasoning**, using connectives such as "Wait a moment", "Wait, does it seem right?", etc. Remember to put your final answer within \boxed{}.
}
\newline

\noindent\underline{Vision-G1: }\texttt{
You FIRST think about the reasoning process as an internal monologue and then provide the final answer. The reasoning process MUST BE enclosed within <think></think> tags. The final answer MUST BE put in \boxed{}.
}
\newline

\noindent\underline{Vision-R1: }\texttt{
A conversation between User and Assistant. The user asks a question, and the Assistant solves it. The assistant first thinks about the reasoning process in the mind and then provides the user with the answer. The reasoning process and answer are enclosed within <think> </think> and <answer> </answer> tags, respectively, i.e., <think> reasoning process here </think><answer> answer here </answer>.
}
\newline

\subsection{Dataset Prompts}

We use prompts from the respective papers for OmniSpatial, MindCube and Spatial457 as recommended.

\noindent\underline{OmniSpatial}: \texttt{
Task \textbackslash n
----- \textbackslash n
You will receive 
1. **Image** - a single RGB frame depicting a scene. \textbackslash n
2. **Question** - a natural-language query about spatial relationships between objects in the image. \textbackslash n
3. **Options** - >=2 answer candidates, each tagged by a capital letter (A, B, C, D…).\textbackslash n
Based on the image and question, provide your answer.
Always ground your answer in the visual evidence; do not hallucinate unseen objects.
If uncertain, pick the most plausible option—never refuse or reply “insufficient information.”
}
\newline

\noindent\underline{MindCube}: \texttt{
Your task is to analyze the spatial arrangement of objects in the scene by examining the provided images, which show the scene from different viewpoints.
}
\newline


\subsection{LLM Judge Scoring Prompts}
\label{app:judge_prompts}
We show the prompts we use for judging MLM generations.\\

\noindent\underline{MCQ Scoring}:\texttt{
You are a helpful assistant. \textbackslash n\textbackslash n
Your task: given (1) a free-form "Response" and (2) a list of "Options",
decide which option the response most likely corresponds to and return the option letter. If no option clearly matches, output "0". \textbackslash n\textbackslash n
Inputs:
- Response: free-form text that may include a letter, a phrase, or an explanation.
- Options: A series of choices, each starting with a single uppercase letter followed by ".", one option in each line.\textbackslash n\textbackslash n
Output format:
- STRICTLY OUTPUT EXACTLY ONE CHARACTER: a single uppercase option letter from the allowed set, or "0".\textbackslash n
- Do not output any explanation, spaces, punctuation, or additional text.\textbackslash n\textbackslash n
Rules:\textbackslash n
1) If the response explicitly names exactly one letter (patterns like "A", "A)", "Option A", "Answer is C"), return that letter immediately.\textbackslash n
2) Only evaluate the explicitly provided choice. If the response is long and complex without an explicit final choice, return "0".\textbackslash n
3) If multiple choices appear in the response, the last unambiguous one is the final choice.\textbackslash n
4) Never judge factual correctness—only map the response to the best matching option letter from the given options.\textbackslash n
5) If no explicit letter can be extracted from the response, compare the response's meaning to option texts. If exactly one option clearly restates or is a synonym/number/name/unit match for the response, return its letter. (Example: response “1956” matches option “B. 1956”)\textbackslash n
6) If the response uses standard MCQ phrases such as "none of the above" or "all of the above" and a matching option exists, map them. If there is no matching option, output "0".\textbackslash n
7) If the response contains both an explicit letter and a conflicting phrase, prefer the explicit letter. If conflicts remain or are unclear, output "0".\textbackslash n
8) If the response says "I don't know", "Cannot determine", or similar, output "0".\textbackslash n
\textbackslash n\textbackslash n
- Example 1\textbackslash n
Response:\textbackslash n
Rome\textbackslash n
Options:\textbackslash n
A. Paris\textbackslash n
B. Berlin\textbackslash n
C. Rome\textbackslash n
D. Madrid\textbackslash n\textbackslash n
Output -> C\textbackslash n\textbackslash n
- Example 2\textbackslash n
Response:\textbackslash n
I don't know\textbackslash n\textbackslash n
Options:\textbackslash n
A. Glucose\textbackslash n
B. Fructose\textbackslash n
C. Sucrose\textbackslash n
D. Lactose\textbackslash n\textbackslash n
Output -> 0\textbackslash n\textbackslash n
- Example 3\textbackslash n
Response:\textbackslash n
A. B\textbackslash n\textbackslash n
Options:\textbackslash n
A. B\textbackslash n
B. D\textbackslash n
C. A\textbackslash n
D. C\textbackslash n\textbackslash n
Output -> A\textbackslash n\textbackslash n
}
\newline

\noindent\underline{VQA Scoring}:\texttt{
You are a helpful assistant.\textbackslash n\textbackslash n
Task: Given a short free-form "Response" and a gold-standard "Gold", decide if the Response expresses the SAME answer as Gold. Output "1" for match, "0" otherwise.\textbackslash n\textbackslash n
Inputs:\textbackslash n
- Gold: the gold-standard answer which is either (i) a short phrase, (ii) an integer, or (iii) "Yes"/"No".\textbackslash n
- Response: a few words or a short phrase, possibly will include reasoning steps before the final answer.\textbackslash n\textbackslash n
Output format:\textbackslash n
- STRICTLY OUTPUT EXACTLY ONE CHARACTER: "1" if matching, "0" if not.\textbackslash n
- Do not output any explanation, spaces, punctuation, or additional text.\textbackslash n\textbackslash n
Rules:\textbackslash n
1) Compare only the final answer in the Response to Gold. Ignore any reasoning steps or intermediate answers present in the Response.\textbackslash n
2) If multiple conflicting answers or uncertainty like "I don't know" appear in the Response, output "0".\textbackslash n
3) Do not use external knowledge; judge only based on the text in Gold and Response.\textbackslash n
4) Punctuation, grammar, and minor spelling errors should be ignored.\textbackslash n
    - uppercase/lowercase differences should be ignored.\textbackslash n
    - hyphen and underscore are ignored. For ex, "double-bus" and "double bus" are considered the same.\textbackslash n
    - synonyms of "Yes"/"No" like "Y"/"N", "True"/"False" must be considered the same.\textbackslash n
    - word representations of numbers like "one"/"two"/"three" must be considered the same as "1"/"2"/"3".\textbackslash n
5) Core concept and critical attributes must match. For example, "New York City" and "New York State" do not match. Other examples of non-matches are  “bus” vs “double bus”; “red” vs “light red”; “dog” vs “golden retriever”; “apple” vs “green apple”.\textbackslash n
6) If the response says "I don't know", "Cannot determine", or similar, output "0".\textbackslash n\textbackslash n
Examples:\textbackslash n
- Gold: Double Bus | Response: This is a bus -> 0\textbackslash n
- Gold: Double Bus | Response: I can see a double-bus -> 1\textbackslash n
- Gold: Yes        | Response: Y -> 1\textbackslash n
- Gold: 10         | Response: ten -> 1\textbackslash n
- Gold: red        | Response: light red -> 0\textbackslash n
- Gold: stop sign  | Response: a stop sign on a pole -> 1\textbackslash n
- Gold: person     | Response: man -> 0\textbackslash n\textbackslash n
Now read the following Gold and Response and output exactly one character: "1" or "0".\textbackslash n
}
\newline

\section{Expanded Tables}
\label{appx:expanded_tables}
Below we present the full table for the No-Img Ablation containing results for all $14$ datasets. 

\begin{table*}[t]
    \centering
    \renewcommand{\arraystretch}{1.12}
    \setlength{\tabcolsep}{4pt}

    \parbox[t]{0.45\textwidth}{
    \centering
    \resizebox{\linewidth}{!}{
    \begin{tabular}{l|cc}
    \toprule
    Benchmark  & \#Questions & Tags\\
    \midrule
    BLINK~\citeyearpar{BLINK} & 1.9K & DEP, REL, CNT, LOC\\
    CV-Bench2D~\citeyearpar{cambrian1} &  1.4K & REL, CNT, LOC, SIZ\\
    MMVP~\citeyearpar{MMVP} & 300 & REL, LOC \\
    RealWorldQA~\citeyearpar{RealWorldQA_xai} & 765 & REL, LOC \\
    SpatialBench~\citeyearpar{SpatialBench} & 174* & REL, LOC, SIZ \\
    VSR~\citeyearpar{VSR} & 1.2K & REL, ORI, EGO\\
    V*Bench~\citeyearpar{VstarBench} & 191 & ATT, REL \\
    \end{tabular}
    }
    }
    \parbox[t]{0.50\textwidth}{
    \centering
    \resizebox{\linewidth}{!}{
    \begin{tabular}{l|cc}
    \toprule
    Benchmark & \#Questions & Tags\\
    \midrule
    3DSRBench~\citeyearpar{ma20243dsrbench} & 5.2K & 3D, LOC, ORI \\
    CV-Bench3D~\citeyearpar{cambrian1} &  1.2K & DEP, 3D, REL\\
    MindCube~\citeyearpar{mindcube}  & 1K & MV, REL, EGO, INT \\
    MMSIBench~\citeyearpar{MMSIBench} &  1K & MV, TMP, LOC, ATT \\
    OmniSpatial~\citeyearpar{omnispatial25} & 1.5K & REL, TMP, INT, EGO\\
    SAT-Real~\citeyearpar{ray2025sat} & 150* & TMP, INT, EGO \\
    \end{tabular}
    }
    }
    \caption{Summary of benchmark datasets used to measure spatial reasoning capabilities of MRMs. Star next to size indicates circular evaluation for those datasets. The tags are REL: object-object spatial relations, DEP: depth/relative distance, ORI: orientation, LOC: localization, SIZ: scale comparison, CNT: counting, 3D: explicit 3D geometry (location \& orientation), MV: multi-image reasoning, TMP: motion/dynamics, EGO: egocentric/allocentric reference, INT: interaction, ATT: object attribute.}
    \label{tab:dataset_summary}
\end{table*}

\begin{table*}[htb]
    \centering
    \scalebox{0.8}{
    \begin{tabular}{l|ccccccc}
    \toprule
    Models & 3DSRBench & BLINK & \multicolumn{2}{c}{CV-Bench} & MindCube & MMSIBench & MMVP \\
    \cmidrule(lr){4-5}&&& 2D & 3D && \\
    \midrule
    Qwen2.5-VL-7B$_{cot}$ & 49.37 & 38.03 & 39.08 & 58.42 & 28.00 & 25.50 & 49.00 \\
    Qwen2.5-VL-7B & 51.10 & 38.51 & 29.14 & 55.92 & 32.10 & 25.70 & 50.00 \\
    \midrule
    \rowcolor{light-gray}
    GThinker-7B & 49.72 & 38.66 & 54.87 & 60.50 & 40.95 & 24.20 & 49.67 \\
    R1-Onevision-7B & 38.20 & 26.67 & 30.53 & 46.58 & 25.81 & 12.60 & 23.67 \\
    \rowcolor{light-gray}
    ViGoRL-7B-Spatial & 47.70 & 39.77 & 55.29 & 65.00 & 39.90 & 26.30 & 48.33 \\
    VL-Rethinker-7B & 51.75 & 40.08 & 36.23 & 59.42 & 34.38 & 24.40 & 49.67 \\
    \rowcolor{light-gray}
    Vision-G1 & 51.85 & 38.66 & 43.60 & 60.17 & 42.57 & 28.90 & 50.00 \\
    Vision-R1-7B & 48.23 & 34.88 & 30.18 & 57.92 & 35.24 & 21.70 & 48.00  \\
    \rowcolor{light-gray}
    TreeVGR-7B & 48.61 & 38.93 & 34.49 & 56.25 & 42.48 & 25.00 & 47.33 \\
    ThinkLite-VL-7B & 51.39 & 40.24 & 34.77 & 60.08 & 32.10 & 26.00 & 50.00 \\
    \end{tabular}
    }
    \vspace{1em}
    \scalebox{0.8}{
    \begin{tabular}{l|cccccc|c}
    \toprule
    Models & OmniSpatial & RealWorldQA & SAT & SpatialBench & VSR & V*Bench & Avg. \\
    \midrule
    Qwen2.5-VL-7B$_{cot}$ & 14.61 & 36.99 & 44.00 & 31.53 & 49.75 & 22.51 & 37.45 \\
    Qwen2.5-VL-7B & 15.26 & 36.73 & 47.67 & 35.63 & 48.85 & 35.08 & 38.59 \\
    \midrule
    \rowcolor{light-gray}
    GThinker-7B & 37.70 & 44.18 & 46.67 & 38.62 & 50.25 & 38.22 & 44.17 \\
    R1-Onevision-7B & 20.81 & 28.37 & 41.00 & 31.53 & 11.78 & 27.75 & 28.10 \\
    \rowcolor{light-gray}
    ViGoRL-7B-Spatial & 24.07 & 40.65 & 49.00 & 36.94 & 48.04 & 40.31 & 43.18 \\
    VL-Rethinker-7B & 21.98 & 42.09 & 51.33 & 37.69 & 50.16 & 37.17 & 41.26 \\
    \rowcolor{light-gray}
    Vision-G1 & 37.90 & 44.05 & 51.00 & 42.72 & 51.47 & 35.08 & 44.46 \\
    Vision-R1-7B & 31.64 & 43.27 & 55.67 & 39.37 & 50.08 & 38.74 & 41.15 \\
    \rowcolor{light-gray}
    TreeVGR-7B & 39.92 & 41.31 & 51.67 & 36.01 & 50.41 & 32.46 & 41.91 \\
    ThinkLite-VL-7B & 32.22 & 43.14 & 55.33 & 38.25 & 51.55 & 37.17 & 42.48 \\
    \end{tabular}
    }
    \caption{Dataset wise expanded results for the No-Image ablation}
    \label{tab:appx_noimg_full}
\end{table*}

In the expanded table below, we provide dataset-wise number for CoT vs Non-CoT performance of MLMs of various backbones and sizes.

\begin{table*}[htb]
    \centering
    \scalebox{0.95}{
    \begin{tabular}{l|ccccccc}
    \toprule
    Models & 3DSRBench & BLINK & \multicolumn{2}{c}{CV-Bench} & MindCube & MMSIBench & MMVP \\
    \cmidrule(lr){4-5}&&& 2D & 3D && \\
    \midrule
    \rowcolor{light-gray}
    Qwen2.5-VL-3B$_{cot}$ & 53.38 & 45.13 & 71.21 & 68.00 & 40.38 & 24.60 & 63.67 \\
    Qwen2.5-VL-3B        & 52.10 & 47.97 & 70.58 & 74.00 & 43.71 & 25.80 & 64.67 \\
    \rowcolor{light-gray}
    Qwen2.5-VL-7B$_{cot}$ & 57.11& 53.44& 75.92& 76.09& 30.83& 27.47& 72.44 \\
    Qwen2.5-VL-7B & 55.38& \textbf{56.04}& 77.17& 83.78& 35.11& 26.87& 75.78 \\
    \rowcolor{light-gray}
    Qwen2.5-VL-72B$_{cot}$ & 61.66 & 58.50 & 78.86 & 85.42 & 39.90 & 29.50 & 76.33 \\
    Qwen2.5-VL-72B        & 59.74 & 63.07 & 79.90 & 86.00 & 42.48 & 32.90 & 79.00 \\
    \midrule
    \rowcolor{light-gray}
    InternVL3-8B$_{cot}$ & 47.93 & 37.66 & 44.92 & 50.42 & 41.62 & 23.60 & 49.67 \\
    InternVL3-8B        & 51.19 & 38.82 & 45.06 & 57.08 & 36.57 & 28.10 & 49.67 \\
    \rowcolor{light-gray}
    InternVL3.5-38B$_{cot}$ & 56.10 & 60.23 & 82.06 & 90.17 & 31.43 & 12.70 & 81.00 \\
    InternVL3.5-38B        & 59.80 & 64.49 & 81.99 & 87.58 & 47.05 & 30.60 & 81.33 \\
    \midrule
    \rowcolor{light-gray}
    LLaVA-1.6-7B$_{cot}$ & 45.01 & 30.04 & 37.41 & 47.75 & 40.10 & 26.50 & 45.00 \\
    LLaVA-1.6-7B        & 45.55 & 21.20 & 41.03 & 54.50 & 40.29 & 29.50 & 49.33 \\
    \rowcolor{light-gray}
    LLaVA-OV-72B$_{cot}$ & 60.48 & 54.81 & 79.49 & 79.42 & 37.52 & 30.30 & 81.00 \\
    LLaVA-OV-72B        & 60.15 & 58.55 & 80.46 & 85.17 & 48.57 & 30.20 & 84.00 \\
    \midrule
    \rowcolor{light-gray}
    GPT-4o$_{cot}$ & 63.20 & 61.60 & 78.23 & 86.42 & 43.52 & 34.10 & 84.33 \\
    GPT-4o        & 61.80 & 65.23 & 75.17 & 85.42 & 47.24 & 34.20 & 84.33 \\
    \end{tabular}
    }
    \vspace{1em}
    \scalebox{0.9}{
    \begin{tabular}{l|ccccccc}
    \toprule
    Models & OmniSpatial & RealWorldQA & SAT & SpatialBench & VSR & V*Bench & Avg. \\
    \midrule
    \rowcolor{light-gray}
    Qwen2.5-VL-3B$_{cot}$ & 40.77 & 62.35 & 55.00 & 56.72 & 74.88 & 71.20 & 55.95 \\
    Qwen2.5-VL-3B        & 45.92 & 65.88 & 59.00 & 56.53 & 79.21 & 75.39 & 58.52 \\
    \rowcolor{light-gray}
    Qwen2.5-VL-7B$_{cot}$ & 40.40& 63.05& 59.22& 61.75& 81.83& 76.27& 59.68 \\
    Qwen2.5-VL-7B & 45.23& 69.02& 63.11& 62.87& 85.38 & 79.06 & 62.68 \\
    \rowcolor{light-gray}
    Qwen2.5-VL-72B$_{cot}$ & 47.68 & 71.76 & 67.33 & 68.66 & 85.35 & 67.54 & 64.50 \\
    Qwen2.5-VL-72B        & 49.51 & 73.73 & 71.00 & 69.40 & 87.64 & 78.01 & 67.11 \\
    \midrule
    \rowcolor{light-gray}
    InternVL3-8B$_{cot}$ & 36.92 & 44.05 & 47.00 & 44.22 & 50.57 & 30.37 & 42.23 \\
    InternVL3-8B        & 36.07 & 40.39 & 46.33 & 43.66 & 49.51 & 34.03 & 42.81 \\
    \rowcolor{light-gray}
    InternVL3.5-38B$_{cot}$ & 48.08 & 69.80 & 64.33 & 61.38 & 81.10 & 66.49 & 61.91 \\
    InternVL3.5-38B        & 48.27 & 76.21 & 64.67 & 68.10 & 83.88 & 69.11 & 66.39 \\
    \midrule
    \rowcolor{light-gray}
    LLaVA-1.6-7B$_{cot}$ & 28.70 & 32.29 & 43.00 & 36.38 & 45.50 & 25.65 & 37.18 \\
    LLaVA-1.6-7B        & 22.50 & 39.35 & 35.00 & 36.19 & 49.26 & 34.55 & 38.33 \\
    \rowcolor{light-gray}
    LLaVA-OV-72B$_{cot}$ & 43.77 & 69.02 & 61.00 & 66.42 & 78.07 & 68.06 & 62.26 \\
    LLaVA-OV-72B        & 48.27 & 71.76 & 66.00 & 69.22 & 79.71 & 67.54 & 65.35 \\
    \midrule
    \rowcolor{light-gray}
    GPT-4o$_{cot}$ & 45.73 & 73.59 & 68.67 & 63.99 & 84.45 & 64.40 & 65.56 \\
    GPT-4o        & 46.44 & 76.60 & 64.33 & 63.25 & 80.93 & 60.73 & 65.05 \\
    \end{tabular}
    }
    \caption{Dataset wise table for averages shown in Figure~\ref{fig:cot_vs_noncot_mlm_bar}.}
    \label{tab:appx_full}
\end{table*}

\noindent\textbf{Gthinker-7B:} visual clues (description of salient part in text) within \textless vcues\_*\textgreater \textless/vcues\_*\textgreater tags, encourage rethinking (to enable reflection \& relook). Build 7k sample CoT dataset using `pattern-guided cold start' (from ScienceQA, M3CoT, Math, Sherlock etc). CoT data generated using a cascade of MLMs ensuring some samples have rethinking stages. RL data sources are very diverse and general (from llava-o1, r1onevision, mm-eureka). 4K samples are pick from RL sources post clustering to enforce diversity. RL \& SFT data sources are different.

\noindent\textbf{ViGoRL-7B:} base models do not perform visual verification \& don't perform backtracking/reflection. Vanilla GRPO also does not incentivize this behavior. Two step process, i) warm-start CoT SFT: MCTS to generate grounded reasoning steps, where each reasoning step anchors though to image coordinates $<s_t, (x_t, y_t)>$. Pref MCTS over linear rollouts to enforce exploration and corrective reflection. Qwen2.5-VL-72B teacher is used to generate about 20k reasoning traces from 1400 images of SAT data (from a total of 32k images) ii) spatially grounded RL: Entire training set of 32k SAT questions. 

\noindent\textbf{Vision-G1:} multi-domain data curation. Training data spanning many domains (all domains of VisCoT) is collected. The training sources have cross image reasoning data IconQA, NLVR2, ImageCode and spatial reasoning datasets VQA-AS, Super-CLEVR. Multi-round RL with data curriculum is used (i.e after every round of RL data selection is performed to discard low quality data). Influence function based selection is done using LESS (LESS: Selecting Influential Data for Targeted Instruction Tuning). For difficulty based filtering, use prev round checkpoint to generate $k$ rollouts for each sample, and retain those with avg. acc between $0.2 \& 0.8$ (dicard too easy and too hard samples). First IF selection gives 40K training samples, then prev round checkpt is used to perform difficulty-based filtering, on which current round RL training is done. Training done for 3 rounds. Unclear if selection \& filtering is done fresh over entire data for every round, or selection done once and filtering done on same set of 40k samples. 

\noindent\textbf{Vision-R1:} Two stage: CoT SFT on 200k `cold-start initialization data' (from llava-cot \& mulberry) Vision-R1-Cold followed by GRPO training on 10k RL data (WeMath, MathVision, PolyMath, SceMQA, Geometry3K) using a Progressive Thinking Suppression Training (PTST) strategy. They observe simple R1-zero style training fails (reasons could be lack of question coverage/diversity, no difficulty based filtering). TO generate CoT data, they perform `Modality-Bridging' by getting text descriptions of images, and feeding it to Deepseek-R1. PTST keeps an output length constraint of $4K, 8K, 16K$ tokens for three stages (each 100 iters), with group sizes $16, 8, 4$ resp. This curriculum enforces shorter reasoning chains at the beginning of GRPO training.

\begin{table*}[htb]
    \centering
    \small
    \scalebox{0.9}{
    \begin{tabular}{l|cccccc}
    \toprule
    method & SFT & RL data & RL Algo & Rewards & Training & Other \\
    \midrule
    GThinker-7B  & CoT-SFT on & diversity based&DAPO & fmt + acc & SFT 3 epoch & tab. 3\&4 ablation  \\
    &7K dset&4K samples& & Hybrid & RL 170 steps&rethink drops gen perf\\
    \midrule
    ViGoRL-7B & MCTS CoT SFT& full train set& GRPO& format + acc & SFT 3 ep, RL 500& Multi-turn RL \\
    &on train subset&&&+coord fmt& steps, kl\_coef 0.01& for vis. search \\
    \midrule
    Vision-G1 &no SFT & IF selection (40K),& GRPO& fmt + acc& 3-round RL&multi-round RL\\
    &&difficulty filtering&w/o KL&& 25 ep each round& \\
    \midrule
    Vision-R1 &CoT SFT 200K &10K no&GRPO&fmt + acc& SFT 2 ep, RL& two above show\\
    &Modality-bridging&filtering&&&300 iters PTST& cold-RFT possible\\
    \end{tabular}}
    \caption{Various methodological aspects of baselines}
    \label{tab:methods}
\end{table*}

\begin{table}[t]
    \centering
    \small
    \begin{tabular}{l|cc}
    \toprule
    Benchmark & CoT & Non-CoT \\
    \midrule
    3DSRBench & 60.67 & 59.69 \\
    BLINK & 59.13 & \textbf{66.70} \\
    CV-Bench2D & 78.65 & 79.21 \\
    CV-Bench3D & \textbf{92.75} & 92.67 \\
    MindCube & 35.14 & 35.14 \\
    MMSIBench & 28.70 & \textbf{30.40} \\
    MMVP & 76.67 & \textbf{79.33} \\
    OmniSpatial & 40.90 & \textbf{45.73} \\
    RealWorldQA & \textbf{73.73} & 70.98 \\
    SAT & \textbf{74.00} & 70.33 \\
    SpatialBench & \textbf{67.91} & 62.87 \\
    VSR & 82.82 & \textbf{84.21} \\
    V*Bench & 79.58 & \textbf{81.68} \\
    \midrule
    Average & 65.43 & \textbf{66.07} \small{$(+0.64)$} \\
    \bottomrule
    \end{tabular}
    \caption{Dataset-wise CoT vs Non-CoT results for Qwen3-VL-8B-Thinking~\cite{qwen3vl}. Non-CoT outperforms CoT on 8 of 13 datasets ($+0.64\%$ avg).}
    \label{tab:qwen3vl_results}
\end{table}

\section{Qualitative Examples}
\label{app:qualitative}

Below we present additional qualitative examples illustrating the failure modes identified in our analysis.

\subsection{GThinker Degenerate Output}

When evaluated under the non-CoT (direct answer) prompt, GThinker fails to follow the direct-answer format and produces degenerate outputs for many samples. An example from RealWorldQA is shown below:

\noindent\textit{Question:} ``Which item is larger than the other?''

\noindent\textit{GThinker output:} \texttt{<tool\_call>\textbackslash n\textbackslash n\textbackslash n\textbackslash n\textbackslash n\textbackslash n\textbackslash n\textbackslash n...} (repeats until max token limit)

This degenerate behavior explains the $-23.14\%$ drop reported for GThinker in Figure~\ref{fig:cot_vs_noncot_mlm_bar} (left table).

\subsection{No-Image++ Hallucination}

In the No-Image++ setting, the input image is replaced with a blank gray image and a ``Cannot determine from the image'' option is added. Despite receiving no visual information, CoT-prompted models hallucinate detailed spatial reasoning. An example from CVBench2D:

\noindent\textit{Question:} ``Where is the cave located with respect to the trees?''

\noindent\textit{Input:} Blank gray image + options including ``Cannot determine from the image''

\noindent\textit{ViGoRL response:} \texttt{<think>} \textit{The cave entrance appears to be at the bottom of the image, while the trees are higher up, suggesting that the cave entrance is below the trees. (300, 450) To determine the position of the cave entrance relative to the trees, I need to consider their vertical alignment. The cave entrance seems to be positioned lower in the image compared to the trees...} \texttt{</think> <answer>} below \texttt{</answer>}

The model hallucinates spatial coordinates and reasons using world knowledge (caves are typically located below trees) rather than acknowledging the absence of visual information.


\end{document}